# Do All Fragments Count?

**Rens Bod**

Informatics Research Institute

University of Leeds, Leeds LS2 9JT, UK

and

Institute for Logic, Language and Computation

University of Amsterdam, Spuistraat 134, 1012 VB, NL

rens@comp.leeds.ac.uk

*We aim at finding the minimal set of fragments which achieves maximal parse accuracy in Data Oriented Parsing. Experiments with the Penn Wall Street Journal treebank show that counts of almost arbitrary fragments within parse trees are important, leading to improved parse accuracy over previous models tested on this treebank. We isolate a number of dependency relations which previous models neglect but which contribute to higher parse accuracy.*

## 1. Introduction

One of the main goals in statistical natural language parsing is to find the minimal set of statistical dependencies (between words and syntactic structures) which achieves maximal parse accuracy. Many stochastic parsing models use linguistic intuitions to find this minimal set, for example by restricting the statistical dependencies to the locality of headwords of constituents (Collins 1997, 1999; Eisner 1997), leaving it as an open question whether there exist important statistical dependencies that go beyond linguistically motivated dependencies. The Data Oriented Parsing model, on the other hand, takes a rather extreme view on this issue: it does not single out a narrowly predefined set of structures as the statistically significant ones; given an annotated corpus, all fragments (i.e., subtrees) seen in that corpus, regardless of size and lexicalization, are in principle taken to form a grammar (see Bod 1992, 1998; Bod & Kaplan 1998; Bonnema et al. 1997; Cormons 1999; Goodman 1996, 1998; Kaplan 1996; de Pauw 2000; Scha 1990; Sima'an 1995, 1999; Way 1999). The set of subtrees that is used is thus very large and extremely redundant. Both from a theoretical and from a computational perspective we may wonder whether it is possible to impose constraints on the subtrees that are used, in such a way that the accuracy of the model does not deteriorate or perhaps even improves. That is the main question addressed in this paper. We report on experiments carried out with the Penn Wall Street Journal (WSJ) treebank to investigate several strategies for constraining the set of subtrees. We found that the only constraints that do not decrease the parse accuracy consist in an upper bound of the number of words in the subtree frontiers and an upper bound on the depth of unlexicalized subtrees. We also found that counts of subtrees with nonheadwords are important, resulting in improved parse accuracy over previous models tested on the WSJ: 90.8% labeled precision and 90.6% labeled recall for sentences ≤ 40 words, and 89.7% labeled precision and recall for sentences ≤ 100 words.

The rest of this paper is organized as follows. We first explain the original Data Oriented Parsing model, known as DOP1, and go into some of its computational properties. We then report on a series of experiments carried out with this model on the Penn Wall Street Journal treebank. We conclude with a discussion of our results and of directly related work.

## 2. The DOP1 model

To-date, the Data Oriented Parsing model has mainly been applied to corpora of trees whose labels consist of primitive symbols (but see Bod et al. 1996; Bod & Kaplan 1998; Cormons 1999 and Way 1999 for more sophisticated DOP models). Let us illustrate the original DOP model presented in Bod (1992), called DOP1, with a simple example. Assume a corpus consisting of only two trees:

(1)

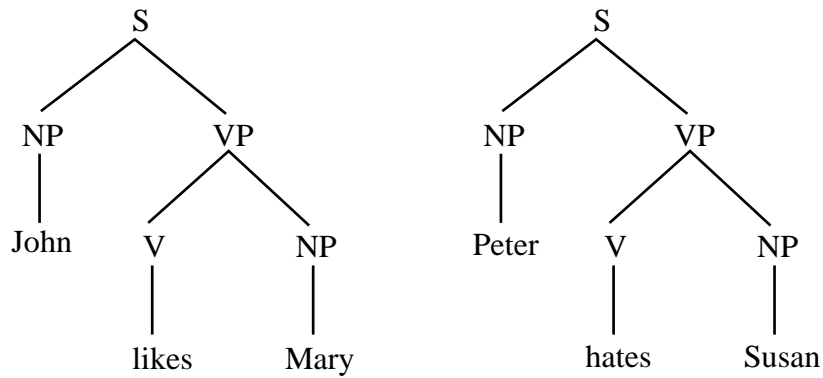

New sentences may be derived by combining fragments, i.e. subtrees, from this corpus, by means of a node-substitution operation indicated as ∘. Node-substitution identifies the leftmost nonterminal frontier node of one subtree with the root node of a second subtree (i.e., the second subtree is *substituted* on the leftmost nonterminal frontier node of the first subtree). Under the convention that the node-substitution operation is left-associative, a new sentence such as *Mary likes Susan* can be derived by combining subtrees from this corpus, as in figure (2):

(2)

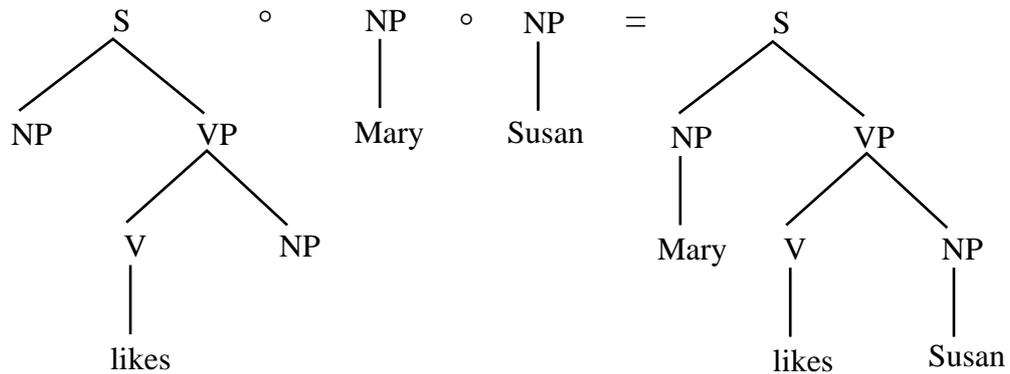

Other derivations may yield the same parse tree, as in figures (3) and (4):

(3)

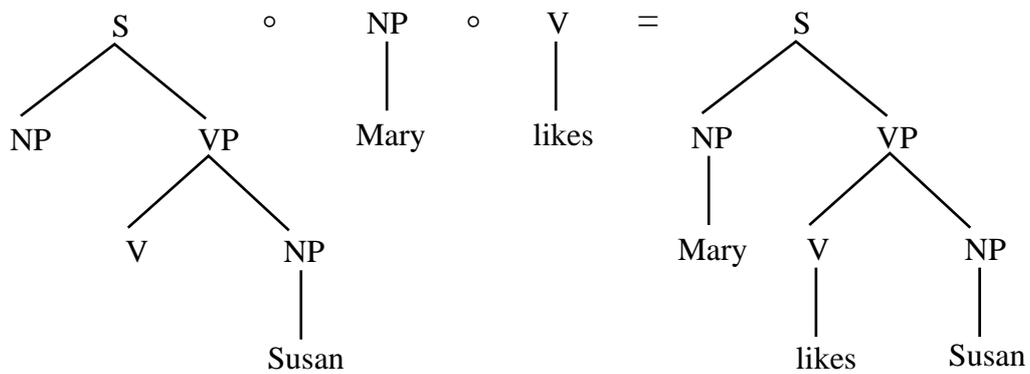

(4)

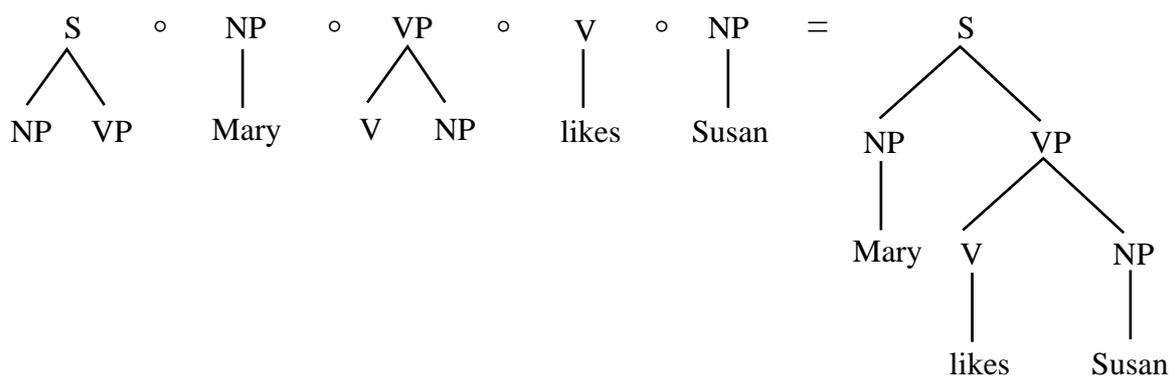

Note that the derivation in (4) corresponds to a context-free derivation in that each of its subtrees is equivalent to a simple context-free rule.

DOP1 computes the probability of a subtree *t* as the probability of selecting *t* among all corpus subtrees that can be substituted on the same node as *t*. This probability is equal to the number of occurrences of *t*, | *t* |, divided by the total number of occurrences of all subtrees *t'* with the same root label as *t*. Let *r(t)* return the root label of *t*. Then we may write:

$$P(t) = \frac{|t|}{\sum_{t': r(t')=r(t)} |t'|}$$

The probability of a derivation $t_1 \circ ... \circ t_n$ is computed by the product of the probabilities of its subtrees $t_i$:

$$P(t_1 \circ ... \circ t_n) = \prod_i P(t_i)$$

As we have seen, there may be several distinct derivations that generate the same parse tree. The probability of a parse tree *T* is thus the sum of the probabilities of its distinct derivations. Let $t_{id}$ be the *i*-th subtree in the derivation *d* that produces tree *T*, then the probability of *T* is given by

$$P(T) = \sum_d \prod_i P(t_{id})$$

Thus DOP1's tree probability model corresponds to a *sum-of-products* model (van Santen 1993) in that it computes the probability of a tree as a sum of products, where each product corresponds to the probability of a distinct derivation generating the particular tree. This distinguishes DOP1 from most other statistical parsing models that identify exactly one derivation for each parse tree and thus compute the probability of a tree by only one product of probabilities -- see Collins (1997, 1999), Charniak (1997, 2000) and Eisner (1997). DOP1's probability model allows for considering counts of subtrees of a wide range of sizes: everything from counts of single-level rules to counts of entire trees. This means that the model is sensitive to the frequency of large subtrees while taking into account the smoothing effects of counts of small subtrees. While most other statistical parsing models also back-off fragments by decomposing them into smaller parts (see e.g. Collins 1997; Charniak 1997; Chiang 2000), DOP1's sum-of-products model takes into account fragments of *any* size as it sums up over all possible derivations that generate a particular tree.

Note that DOP1's parameters (i.e. the subtree probabilities) are directly estimated from their simple relative frequencies in the corpus. While the relative frequency estimator obtains very competitive results on several domains (Bod 2000a), it does not maximize the likelihood of the training data; this is because there may be hidden derivations which the relative frequency estimator cannot deal with.[1] However, there are estimation procedures that do take into account hidden derivations and that maximize the likelihood of the training data. For example, Bod (2000b) experimented with a DOP1 model which trains the subtree probabilities by a maximum likelihood reestimation procedure belonging to the class of expectation-maximization algorithms (Dempster et al. 1977), but this resulted in a decrease in parse accuracy on the ATIS and OVIS corpora (Bod 2000a), although it slightly improved the word error rate for OVIS word-graphs. Other ways for estimating DOP1's parameters have also been proposed: Bonnema et al. (1999) estimate the probability of a subtree as the probability that it has been involved in the derivation of a corpus tree, but it is not yet known how their estimator compares experimentally to DOP1's relative frequency estimator. Since the relative frequency estimator has not been surpassed by any other estimator for DOP1 (at least in natural language parsing), we will stick to this estimator for the rest of this paper.

## 3. Computational issues

Bod (1993) showed how standard chart parsing techniques can be applied to DOP1. Each corpus-subtree *t* is converted into a context-free rule *r* where the lefthand side of *r* corresponds to the root label of *t* and the righthand side of *r* corresponds to the frontier labels of *t*. Indices link the rules to the original subtrees so as to maintain the subtree's internal structure and probability. These rules are used to create a derivation forest for a sentence, and the most probable parse is computed by sampling a sufficiently large number of random derivations from the forest ("Monte Carlo disambiguation", see Bod 1998; Chappelier & Rajman 2000). While this technique has been successfully applied to parsing the ATIS portion in the Penn Treebank (Marcus et al. 1993), it is extremely time consuming. This is mainly because the number of random derivations that should be sampled to reliably estimate the most probable parse increases exponentially with the sentence length (see Goodman 1998; Scha et al. 1999). It is therefore questionable whether Bod's sampling technique can be scaled to larger domains such as the WSJ portion in the Penn Treebank.

Goodman (1996, 1998) showed how DOP1 can be reduced to a compact stochastic context-free grammar (SCFG) which contains exactly eight SCFG rules for each node in the training set trees. Although Goodman's method does still not allow for an efficient computation of the most probable parse (in fact, the problem of computing the most probable parse is NP-hard -- see Sima'an 1996), his method does allow for an efficient computation of the "maximum constituents parse", i.e., the parse tree that is most likely to have the largest number of correct constituents (also called the "labeled recall parse"). Goodman has shown on the ATIS corpus that the maximum constituents parse performs at least as well as the most probable parse if all subtrees are used. Unfortunately, Goodman's reduction method only works if indeed *all* subtrees are used, which means that we cannot employ his method to study the effect of various subtree restrictions. Sima'an (1999: 108) argues that there may still be an isomorphic SCFG for DOP1 if the corpus-subtrees are restricted in size or lexicalization, but that the number of the SCFG rules explodes in that case.

---

[1] Only if the subtrees are restricted to depth 1 does the relative frequency estimator coincide with the maximum likelihood estimator. Such a depth-1 DOP model corresponds to a stochastic context-free grammar and is equivalent to the "corpus parser" in Atwell (1988); it was first tested in Bod (1993) and more extensively in Bod (1995); it was next used by Charniak (1996) who called it a "treebank grammar". (NB: DOP models that allow subtrees of greater depth outperform depth-1 DOP models.)

In this paper we will use Bod's subtree-to-rule conversion method for studying the impact of various subtree restrictions on the WSJ corpus. However, we will not use Bod's Monte Carlo sampling technique from complete derivation forests, as this turned out to be prohibitive for WSJ sentences. Instead, we use a Viterbi *n*-best search and estimate the most probable parse from the 1,000 most probable derivations, summing up the probabilities of derivations that generate the same tree. Although this heuristic does not guarantee that the most probable parse is actually found, it is shown in Bod (2000a) to perform at least as well as the estimation of the most probable parse with Monte Carlo techniques. The algorithm for computing the Viterbi *n* most probable derivations follows straightforwardly from the algorithm which computes the most probable derivation by means of a CKY parser (e.g. Jelinek et al. 1990). However, in computing the Viterbi *n* most probable derivations it is prohibitive to keep track of all subderivations at each edge in the chart (at least for such a large corpus as the WSJ). As in most other statistical parsing systems, we therefore use the pruning technique described in Goodman (1997) and Collins (1999: 263-264), which computes the score of an item in the chart by multiplying it by the prior probability of the item (using a bottom-up CKY parser). Any item with a score less than $10^{-5}$ times of that of the best item is pruned from the chart.

## 4. What is the minimal subtree set that achieves maximal parse accuracy?
### 4.1 The base line
For our base line parse accuracy, we used the now standard division of the WSJ (see Collins 1997, 1999; Charniak 1997, 2000; Ratnaparkhi 1999) with sections 2 through 21 for training (approx. 40,000 sentences) and section 23 for testing (2416 sentences ≤ 100 words); section 22 was used as development set. All trees were stripped off their semantic tags, co-reference information and quotation marks. We used all training set subtrees of depth 1, but due to memory limitations we used a subset of the subtrees larger than depth 1, by taking for each depth a random sample of 400,000 subtrees. This random subtree sample was not selected from an exhaustive computation of all subtrees of a particular depth (which is prohibitive for the large WSJ). Instead, for each particular depth > 1 we sampled subtrees by first randomly selecting a node in a random tree from the training set, after which we selected random expansions from that node until a subtree of the particular depth was obtained. We repeated this procedure 400,000 times for each depth $> 1$ and $\leq 14$. Thus no subtrees of depth $> 14$ were used. This resulted in a total set of 5,217,529 subtrees which we will call the "base line subtree set" and which was smoothed by the technique described in Bod (1996) based on Good-Turing. Since our subtrees are allowed to be lexicalized (at their frontiers), we did not use a separate part-of-speech tagger: the test sentences were directly parsed by the training set subtrees. For words that were unknown in our subtree set, we guessed their categories by means of the method described in Weischedel et al. (1993) which uses statistics on word-endings, hyphenation and capitalization. The guessed category for each unknown word was converted into a depth-1 subtree and assigned a probability by means of simple Good-Turing estimation (see Bod 1996). The most probable parse for each test sentence was estimated from the 1,000 most probable derivations of that sentence, as described in section 3. All experiments were carried out on a Sun UltraSPARC 60 with a 450 MHz UltraSPARC-II processor and 2 gigabytes of real memory.

We used the standard PARSEVAL scores (Black et al. 1991) to compare a proposed parse $P$ (i.e. our estimated most probable parse) with the corresponding correct treebank parse $T$ as follows:

$$\text{Labeled Precision} = \frac{\text{\# correct constituents in } P}{\text{\# constituents in } P}$$

$$\text{Labeled Recall} = \frac{\text{\# correct constituents in } P}{\text{\# constituents in } T}$$

A constituent in $P$ is "correct" if there exists a constituent in $T$ of the same label that spans the same words. We used the "evalb" program written by Satoshi Sekine to compute these scores.[2] Although evalb also computes some other scores, such as CB (average number of crossing brackets per sentence), 0CB (percentage of sentences with 0 crossing brackets) etc., we will focus on the Labeled Precision (LP) and Labeled Recall (LR) scores only in this paper, since these scores are commonly used to rank competing parsing systems.

Table 1 shows the LP and LR scores obtained with our base line subtree set, and compares these scores with those of previous stochastic parsers tested on the WSJ (respectively Charniak 1997, Collins 1999, Ratnaparkhi 1999, and Charniak 2000). The average CPU time was approximately 220 seconds per sentence (for sentences $\leq 100$ words).

---

[2] http://www.cs.nyu.edu/cs/projects/proteus/evalb/

| Parser | LP | LR |
|---|---|---|
| | ≤ 40 words | |
| Char97 | 87.4 | 87.5 |
| Coll99 | 88.7 | 88.5 |
| Char00 | 90.1 | 90.1 |
| Bod00 | 89.5 | 89.3 |
| | ≤ 100 words | |
| Char97 | 86.6 | 86.7 |
| Coll99 | 88.3 | 88.1 |
| Ratna99 | 87.5 | 86.3 |
| Char00 | 89.5 | 89.6 |
| Bod00 | 88.6 | 88.3 |

Table 1: Parsing results with the base line subtree set compared to previous systems

The table shows that by using the base line subtree set, our parser outperforms most previous parsers but it performs worse than the parser in Charniak (2000). We will use our scores of 89.5% LP and 89.3% LR (for test sentences ≤ 40 words) as the base line result against which the effect of various subtree restrictions is compared. We will see that while most subtree restrictions diminish the accuracy scores, there are restrictions that improve our scores, even beyond those of Charniak (2000). We will initially study our subtree restrictions only for test sentences ≤ 40 words (2245 sentences), after which we will give in 4.6 our results for all test sentences ≤ 100 words (2416 sentences). While we have initially performed all subtree restrictions on the development set (section 22 from the WSJ), we believe that it is interesting and instructive to also report these subtree restrictions on the test set (section 23) rather than reporting our best result only.

**4.2 The impact of subtree size**

Our first subtree restriction is concerned with subtree size. It may be evident that large subtrees can capture more lexical/structural dependencies than small ones. We are interested in how much these dependencies actually lead to better predictions for the correct parse. Therefore we performed experiments with versions of DOP1 where the subtree set is restricted to subtrees with a certain maximum depth. For instance, restricting the maximum depth of the subtrees to 1 gives us subtrees that cover exactly one level of constituent structure, which makes DOP1 equivalent to a stochastic context-free grammar. For a maximal subtree depth of 2, we obtain subtrees that also cover two levels of constituent structure, which capture some more lexical/syntactic dependencies, etc. The following table shows the results of these experiments.

| depth of subtrees | LP | LR |
|---|---|---|
| 1 | 76.0 | 71.8 |
| ≤2 | 80.1 | 76.5 |
| ≤3 | 82.8 | 80.9 |
| ≤4 | 84.7 | 84.1 |
| ≤5 | 85.5 | 84.9 |
| ≤6 | 86.2 | 86.0 |
| ≤8 | 87.9 | 87.1 |
| ≤10 | 88.6 | 88.0 |
| ≤12 | 89.1 | 88.8 |
| ≤14 | 89.5 | 89.3 |

Table 2. Parsing results for different subtree depths (for test sentences ≤ 40 words)

Our scores for subtree-depth 1 are comparable to Charniak's treebank grammar if tested on word strings (see Charniak 1997)[3]. Our scores are slightly better, which may be due to the use of a different unknown word model. Note that the scores consistently improve if larger subtrees are taken into account. The highest scores are obtained if the full base line subtree set is used, but they remain behind the results of Charniak (2000). One might expect that our results further increase if even larger subtrees are used; but due to memory limitations we did not perform experiments with subtrees larger than depth 14. Bod (1998) conjectures that the increase in parse accuracy with increasing subtree size (where the accuracy increase itself decreases) is independent of the language and the nature of the linguistic representations, and refers to it as the "DOP hypothesis". This hypothesis has also been validated for Lexical-Functional Grammar representations (Bod 2000c).

**4.3 The impact of lexical context**
The more words a subtree contains in its frontier, the more lexical dependencies can be taken into account. To test the impact of the lexical context on the accuracy, we performed experiments with different versions of the model where the base line subtree set is restricted to subtrees whose frontiers contain a certain maximum number of words; the subtree depth in the base line subtree set was not constrained (though no subtrees deeper then 14 were in this base line set). The following table shows the results of our experiments.

---

[3] Charniak (1996) tests his treebank grammar on part-of-speech strings, obtaining somewhat higher LP/LR scores. In the current paper, however, we will only present results on word strings.

| # words in subtrees | LP | LR |
|---|---|---|
| ≤1 | 84.4 | 84.0 |
| ≤2 | 85.2 | 84.9 |
| ≤3 | 86.6 | 86.3 |
| ≤4 | 87.6 | 87.4 |
| ≤6 | 88.0 | 87.9 |
| ≤8 | 89.2 | 89.1 |
| ≤10 | 90.2 | 90.1 |
| ≤11 | 90.8 | 90.4 |
| ≤12 | 90.8 | 90.5 |
| ≤13 | 90.4 | 90.3 |
| ≤14 | 90.3 | 90.3 |
| ≤16 | 89.9 | 89.8 |
| unrestricted | 89.5 | 89.3 |

Table 3. Parsing results for different subtree lexicalizations (for test sentences ≤ 40 words)

The table shows that the accuracy initially increases when the lexical context is enlarged, but that the accuracy decreases if the number of words in the subtree frontiers exceeds 12 words. Our highest scores of 90.8% LP and 90.5% LR outperform the scores of the best previously published parser by Charniak (2000) who obtains 90.1% for both LP and LR. Our scores also outperform the *reranking* technique of Collins (2000) who reranks the output of the parser of Collins (1999) using a boosting method based on Schapire & Singer (1998), and who obtains 90.4% LP and 90.1% LR. We have thus found a subtree restriction which does not decrease the parse accuracy but even improves it. This restriction consists of an upper bound of 12 words in the subtree frontiers, for subtrees ≤ depth 14. (We have also tested this lexical restriction in combination with subtrees smaller than depth 14, but this led to a decrease in accuracy.)

### 4.4 The impact of structural context

Instead of investigating the impact of lexical context we may also be interested in studying the importance of structural context. We may raise the question as to whether we need all *un*lexicalized subtrees, since such subtrees do not contain any lexical information, although they may be useful to smooth lexicalized subtrees. We accomplished a set of experiments where unlexicalized subtrees of a certain minimal depth are deleted from the base line subtree set, while all lexicalized subtrees up to 12 words are retained:

| depth of deleted unlexicalized subtrees | LP | LR |
|---|---|---|
| ≥1 | 79.9 | 77.7 |
| ≥2 | 86.4 | 86.1 |
| ≥3 | 89.9 | 89.5 |
| ≥4 | 90.6 | 90.2 |
| ≥5 | 90.7 | 90.6 |
| ≥6 | 90.8 | 90.6 |
| ≥7 | 90.8 | 90.5 |
| ≥8 | 90.8 | 90.5 |
| ≥10 | 90.8 | 90.5 |
| ≥12 | 90.8 | 90.5 |

Table 4. Parsing results for different structural context (for test sentences ≤ 40 words)

The table shows that the accuracy increases if unlexicalized subtrees are retained, but that unlexicalized subtrees larger than depth 6 do not contribute to any further increase in accuracy. On the contrary, these larger subtrees even slightly decrease the accuracy. The highest scores obtained are: 90.8% labeled precision and 90.6% labeled recall. We thus conclude that pure structural context without any lexical information contributes to higher parse accuracy (even if there exists an upper bound for the size of structural context). This importance of stuctural context is consonant with Johnson (1998) who showed that structural context from higher nodes in the tree (i.e. grand parent nodes) contributes to higher parse accuracy. This mirrors our result of the importance of unlexicalized subtrees of depth 2. But our results show that larger unlexicalized subtrees (up to depth 6) contribute to the parse accuracy as well.

### 4.5 The impact of nonheadword dependencies

We may also raise the question as to whether we need almost arbitrarily large *lexicalized* subtrees (up to 12 words) to obtain our best results. It could be the case that DOP's gain in parse accuracy with increasing subtree depth is due to the model becoming sensitive to the influence of lexical heads higher in the tree, and that this gain could also be achieved by a more compact model which annotates the nonterminals with their headwords. Such "head-lexicalized stochastic grammars" have recently become increasingly popular (e.g. Collins 1997, 1999; Charniak 1997, 2000) and are based on Magerman's head-percolation scheme to determine the headword of each nonterminal (Magerman 1995). However, these head-lexicalized stochastic grammars are not able to capture dependency relations between words that according to Magerman's head-percolation scheme are "nonheadwords" -- e.g. between *more* and *than* in the WSJ construction *carry more people than cargo* where neither *more* nor *than* are headwords of the NP constituent *more people than cargo*. A frontier-lexicalized DOP model, on the other hand, can easily capture these dependencies since it includes subtrees in which *more* and *than* are the only frontier words. One may object that this example is somewhat far-fetched, but Chiang (2000) notes that head-lexicalized stochastic grammars fall short in encoding even simple dependency relations such as between *left* and *John* in the sentence *John should have left*. This is because Magerman's head-percolation scheme makes *should* and *have* the heads of their respective VPs so that there is no dependency relation between the verb *left* and its subject *John*.[4] Chiang observes that almost a quarter of all nonempty subjects in the WSJ appear in such a configuration. It is again trivial to see that a frontier-lexicalized DOP model can capture these dependency relations.

In order to isolate the contribution of nonheadword dependencies to the parse accuracy, we eliminated all subtrees containing a certain maximum number of nonheadwords, where a nonheadword of a subtree is a word which according to Magerman's scheme is not a headword of the subtree's root nonterminal (although such a nonheadword may of course be a headword of one of the subtree's internal nodes). In the following experiments we used the subtree set for which maximum accuracy was obtained in our previous experiments, i.e. containing all lexicalized subtrees with maximally 12 frontier words and all unlexicalized subtrees up to depth 6.

| # nonheadwords in subtrees | LP | LR |
|:---:|:---:|:---:|
| 0 | 89.6 | 89.6 |
| ≤1 | 90.2 | 90.1 |
| ≤2 | 90.4 | 90.2 |
| ≤3 | 90.3 | 90.2 |
| ≤4 | 90.6 | 90.4 |
| ≤5 | 90.6 | 90.6 |
| ≤6 | 90.6 | 90.5 |
| ≤7 | 90.7 | 90.7 |
| ≤8 | 90.8 | 90.6 |
| unrestricted | 90.8 | 90.6 |

Table 5. Parsing results for different number of nonheadwords (for test sentences ≤ 40 words)

The table shows that nonheadwords contribute to higher parse accuracy: the difference between using no and all nonheadwords is 1.2% in LP and 1.0% in LR. Although this difference is relatively small, it does indicate that nonheadword dependencies should not be discarded in the WSJ. We should note, however, that most other stochastic parsers do include counts of *single* nonheadwords: they appear in the backed-off statistics of these parsers (see Collins 1997, 1999; Charniak 1997; Goodman 1998). But our parser is the first parser that also includes counts between two or more nonheadwords, to the best of our knowledge, and these counts lead to improved performance, as can be seen in the table above.

---

[4] Any other head-percolation scheme would have complementary shortcomings.

### 4.6 Results for all test sentences ≤ 100 words

We have seen that for test sentences ≤ 40 words, maximal parse accuracy was obtained by a subtree set which is restricted to subtrees with not more than 12 words and which does not contain unlexicalized subtrees deeper than 6. We used these restrictions to test our model on all sentences ≤ 100 words from the WSJ test set. This resulted in an LP of 89.7% and an LR of 89.7%. These scores slightly outperform the best previously published parser by Charniak (2000), who obtained 89.5% LP and 89.6% LR for test sentences ≤ 100 words. Only the reranking technique proposed by Collins (2000) slightly outperforms our precision score, but not our recall score: 89.9% LP and 89.6% LR.

## 5. Discussion and conclusion

The main goal of this paper was to find the minimal set of fragments which achieves maximal parse accuracy in Data Oriented Parsing. We have found that, as in some electoral systems, not all fragments count. Yet, the minimal set of fragments is still very large and extremely redundant: highest parse accuracy is obtained by employing only two constraints on the fragment set: a restriction of the number of words in the fragment frontiers to 12 and a restriction of the depth of unlexicalized fragments to 6 (resulting in 90.8% LP and 90.6% LR for sentences ≤ 40 words). No other constraints were warranted.

While we understand why certain fragments may be important (such as fragments with certain nonheadwords), we do not understand why maximal parse accuracy occurs with exactly the constraints reported above. We surmise that these constraints simply differ from corpus to corpus and are related to general data sparseness effects. In previous experiments with DOP1 on smaller domains, such as the ATIS and OVIS corpora, we found that the parse accuracy decreases also after a certain maximum subtree depth (see Bod 1998; Bonnema et al. 1997; Sima'an 1999). We expect that also for the WSJ the parse accuracy will decrease after a certain depth, although we have not been able to find this depth so far.

A major difference between our approach and most other models tested on the WSJ is that the DOP model uses frontier lexicalization while most other models use, what we might call, constituent lexicalization (in that it associates each constituent nonterminal with its lexical head -- see Collins 1997, 1999; Charniak 1997; Eisner 1997). The results in this paper, especially those of section 4.5, indicate that our use of frontier lexicalization is a more flexible and promising approach than the use of constituent lexicalization (although our CPU time per sentence is much longer). Our results also show that the linguistically motivated constraint which limits the statistical dependencies to the locality of headwords of constituents is too narrow. Not only are counts of subtrees with nonheadwords important, also counts of unlexicalized subtrees up to depth 6 increase the parse accuracy.

The only other model that uses frontier lexicalization (and that was tested on the standard WSJ split) is Chiang (2000) who extracts a stochastic tree-insertion grammar or STIG (Schabes & Waters 1996) from the WSJ, obtaining 86.6% LP and 86.9% LR for sentences ≤ 40 words. However, Chiang's approach is limited in at least two respects. First, each elementary tree in his STIG is lexicalized with exactly one lexical item, while our results show that there is an increase in parse accuracy if more lexical items are included and also if unlexicalized trees are included (in his conclusion Chiang acknowledges that "multiply anchored trees" may be important). Second, Chiang computes the probability of a tree by taking into account only one derivation, while in STIG, like in DOP1, there can be several derivations that generate the same parse tree. Apart from these shortcomings, we believe that the insertion operation might be a beneficial extension of DOP1, since it allows for enlarging the scope of lexical dependencies (see Hoogweg 2000).

Another difference between our approach and most other models is that the underlying grammar of DOP is based on a treebank grammar (cf. Charniak 1996, 1997), while most current stochastic parsing models use a "markov grammar" (e.g. Collins 1999; Charniak 2000). While a treebank grammar only assigns probabilities to rules or subtrees that are seen in a treebank, a markov grammar assigns probabilities to any possible rule, resulting in a more robust model. We expect that the application of the markov grammar approach to DOP will further improve our results. Research in this direction is already ongoing (see e.g. Sima'an 2000).

Although we believe that our main result is to have shown that *almost arbitrary fragments within parse trees are important*, we find it fascinating that a simple model as DOP1, which was published as early as in 1992, still outperforms all other stochastic parsers in the literature.[5] Yet, DOP is the only model in the literature which does not *a priori* restrict the fragments that may be used to compute the most probable parse of a sentence. Instead, it starts out by taking into account all fragments seen in a treebank and then systematically investigates various fragment restrictions to discover the set of relevant fragments. From this perspective, the DOP approach can be seen as striving for the same goal as other approaches but from a different direction. While other approaches typically limit the statistical dependencies beforehand (for example to headword dependencies) and then try to improve parse accuracy by gradually letting in more dependencies, we start out by taking into account as many dependencies as possible and then try to constrain them without losing parse accuracy. It is not unlikely that these two opposite directions finally converge to the same, true set of statistical dependencies for natural language parsing.[6]

---

[5] Even Goodman (1996, 1998), who was quite critical about our reporting on the ATIS corpus, must concede that DOP1 outperforms Pereira and Schabes (1992) on that corpus.

As it happens, some convergence of these two approaches has already taken place. While earlier head-lexicalized models restricted fragments to the locality of headwords of constituents (e.g. Collins 1996; Eisner 1996), later models showed the importance of including additional context from higher nodes in the tree, resulting in improved parse accuracy (Charniak 1997; Johnson 1998). This mirrors our result of the utility of fragments of depth 2 (and larger) which was already reported in Bod (1993). The importance of including counts of (single) nonheadwords is now also quite uncontroversial (e.g. Collins 1997, 1999; Charniak 2000), and the current paper has shown the importance of including two and more nonheadwords. Recently, Collins (2000) has even observed that "In an ideal situation we would be able to encode arbitrary features $h_s$, thereby keeping track of counts of arbitrary fragments within parse trees, without having to worry about formulating a derivation that included these features.". This philosophy is in perfect correspondence with the DOP approach.

## References

E. Atwell, 1988. Transforming a Parsed Corpus into a Corpus Parser. in: M. Kytö, O. Ihalainen and M. Rissanen (eds.), *Corpus Linguistics, Hard and Soft*, Rodopi, Amsterdam.

E. Black, S. Abney, D. Flickinger, C. Gnadiec, R. Grishman, P. Harrison, D. Hindle, R. Ingria, F. Jelinek, J. Klavans, M. Liberman, M. Marcus, S. Roukos, B. Santorini and T. Strzalkowski, 1991. A Procedure for Quantitatively Comparing the Syntactic Coverage of English, *Proceedings DARPA Speech and Natural Language Workshop*, Pacific Grove, Morgan Kaufmann.

R. Bod, 1992. A Computational Model of Language Performance: Data Oriented Parsing, *Proceedings COLING'92*, Nantes, France.

R. Bod, 1993. Using an Annotated Language Corpus as a Virtual Stochastic Grammar, *Proceedings AAAI'93*, Washington D.C.

---

[6] It should be emphasized that statistical parsers based on phrase-structure trees are intrinsically limited in that there are syntactic and semantic relations in natural language that cannot be encoded in surface trees. An adequate statistical natural language parser should therefore be based on richer linguistic representations, such as LFG representations (e.g. Bod & Kaplan 1998; Bod 2000c; Riezler et al. 2000) or HPSG representations (e.g. Neumann & Flickinger 1999).

R. Bod, 1995. *Enriching Linguistics with Statistics: Performance Models of Natural Language*, ILLC Dissertation Series 1995-14, University of Amsterdam, The Netherlands.

R. Bod, 1996. Two Questions about Data Oriented Parsing, *Proceedings Fourth Workshop on Very Large Corpora*, COLING-1996, Copenhagen, Denmark.

R. Bod, 1998. *Beyond Grammar: An Experience-Based Theory of Language,* CSLI Publications, distributed by Cambridge University Press.

R. Bod, 2000a. Parsing with the Shortest Derivation, *Proceedings COLING'2000*, Saarbrücken, Germany.

R. Bod 2000b. Combining Semantic and Syntactic Structure for Language Modeling, *Proceedings ICSLP-2000*, Beijing, China.

R. Bod 2000c. An Improved Parser for Data-Oriented Lexical-Functional Analysis, *Proceedings ACL'2000*, Hong Kong, China.

R. Bod, R. Bonnema and R. Scha, 1996. A Data-Oriented Approach to Semantic Interpretation, *Proceedings Workshop on Corpus-Oriented Semantic Analysis*, ECAI-96, Budapest, Hungary.

R. Bod and R. Kaplan, 1998. A Probabilistic Corpus-Driven Model for Lexical-Functional Analysis, *Proceedings COLING-ACL'98*, Montreal, Canada.

R. Bonnema, R. Bod and R. Scha, 1997. A DOP Model for Semantic Interpretation, *Proceedings ACL/EACL-97*, Madrid, Spain.

R. Bonnema, P. Buying and R. Scha, 1999. A New Probability Model for Data-Oriented Parsing. *Proceedings of the Amsterdam Colloquium 1999.* Amsterdam, The Netherlands.

J. Chappelier and M. Rajman, 2000. Monte Carlo Sampling for NP-hard Maximization Problems in the Framework of Weighted Parsing, in *Natural Language Processing -- NLP 2000, Lecture Notes in Artificial Intelligence 1835*, D. Christodoulakis (ed.), 2000, 106-117.

E. Charniak, 1996. Tree-bank Grammars, *Proceedings AAAI'96*, Portland, Oregon.

E. Charniak, 1997. Statistical Parsing with a Context-Free Grammar and Word Statistics, *Proceedings AAAI-97*, Menlo Park.

E. Charniak, 2000. A Maximum-Entropy-Inspired Parser. *Proceedings ANLP-NAACL'2000*, Seattle, Washington.

D. Chiang, 2000. Statistical parsing with an automatically extracted tree adjoining grammar, *Proceedings ACL'2000*, Hong Kong, China.

M. Collins, 1996. A new statistical parser based on bigram lexical dependencies, *Proceedings ACL'96*, Santa Cruz (Ca.).

M. Collins, 1997. Three generative lexicalised models for statistical parsing, *Proceedings ACL'97*, Madrid, Spain.

M. Collins, 1999. *Head-Driven Statistical Models for Natural Language Parsing*, PhD-thesis, University of Pennsylvania, PA.

M. Collins, 2000. Discriminative Reranking for Natural Language Parsing, *Proceedings ICML-2000*, Stanford, Ca.

B. Cormons, 1999. *Analyse et desambiguisation: Une approche à base de corpus (Data-Oriented Parsing) pour les répresentations lexicales fonctionnelles*, PhD thesis, Université de Rennes, France.

A. Dempster, N. Laird and D. Rubin, 1977. Maximum Likelihood from Incomplete Data via the EM Algorithm, *Journal of the Royal Statistical Society*, 39:1-38.

J. Eisner, 1996. Three new probabilistic models for dependency parsing: an exploration, *Proceedings COLING-96*, Copenhagen, Denmark.

J. Eisner, 1997. Bilexical Grammars and a Cubic-Time Probabilistic Parser, *Proceedings Fifth International Workshop on Parsing Technologies*, Boston, Mass.

J. Goodman, 1996. Efficient Algorithms for Parsing the DOP Model, *Proceedings Empirical Methods in Natural Language Processing*, Philadelphia, PA.

J. Goodman, 1997. Global Thresholding and Multiple-Pass Parsing, *Proceedings EMNLP-2*, Boston, Mass.

J. Goodman, 1998. *Parsing Inside-Out*, Ph.D. thesis, Harvard University, Mass.

L. Hoogweg, 1999. A Data-Oriented Approach to Tree-Insertion Grammar, *Computational Linguistics in the Netherlands 1999*, Utrecht, The Netherlands.

F. Jelinek, J. Lafferty and R. Mercer, 1990. *Basic Methods of Proba-bilistic Context Free Grammars*, Technical Report IBM RC 16374 (#72684), Yorktown Heights.

M. Johnson, 1998. PCFG Models of Linguistic Tree Representations, *Computational Linguistics* 24(4), 613-632.

R. Kaplan, 1996. A Probabilistic Approach to Lexical-Functional Analysis, *Proceedings of the 1996 LFG Conference and Workshops*, CSLI Publications, Stanford, Ca.

D. Magerman, 1995. Statistical Decision-Tree Models for Parsing, *Proceedings ACL'95*, Cambridge, Mass.

M. Marcus, B. Santorini and M. Marcinkiewicz, 1993. Building a Large Annotated Corpus of English: the Penn Treebank, *Computational Linguistics* 19(2).

G. Neumann and D. Flickinger, 1999. Learning Stochastic Lexicalized Tree Grammars from HPSG, DFKI Technical Report, Saarbrücken, Germany.

G. de Pauw, 2000. Aspects of Pattern-matching in Data-Oriented Parsing, *Proceedings COLING-2000*, Saarbrücken, Germany.

F. Pereira and Y. Schabes, 1992. Inside-Outside Reestimation from Partially Bracketed Corpora, *Proceedings ACL'92*, Newark, Delaware.

A. Ratnaparkhi, 1999. Learning to Parse Natural Language with Maximum Entropy Models, *Machine Learning* 34, 151-176.

S. Riezler, D. Prescher, J. Kuhn and M. Johnson, 2000. Lexicalized Stochastic Modeling of Constraint-Based Grammars using Log-Linear Measures and EM Training, *Proceedings ACL'2000*, Hong Kong, China.

J. van Santen, 1993. Exploring N-way Tables with Sums-of-Products Models. *Journal of Mathematical Psychology* 37, 327-371.

R. Scha, 1990. Taaltheorie en Taaltechnologie; Competence en Performance, in Q.A.M. de Kort and G.L.J. Leerdam (eds.), *Computertoepassingen in de Neerlandistiek*, Almere: Landelijke Vereniging van Neerlandici (LVVN-jaarboek).

R. Scha, R. Bod and K. Sima'an, 1999. Memory-Based Syntactic Analysis, *Journal of Experimental and Theoretical Artificial Intelligence* 11 (Special Issue on Memory-Based Language Processing).

Y. Schabes and R. Waters, 1996. Stochastic Lexicalized Tree-Insertion Grammar. In H. Bunt and M. Tomita (eds.) *Recent Advances in Parsing Technology*. Kluwer Academic Publishers.

R. Schapire and Y. Singer, 1998. Improved Boosting Algorithms Using Confedence-Rated Predictions, *Proceedings 11th Annual Conference on Computational Learning Theory*. Morgan Kaufmann, San Francisco.

K. Sima'an, 1995. An optimized algorithm for Data Oriented Parsing, *Proceedings International Conference on Recent Advances in Natural Language Processing*, Tzigov Chark, Bulgaria.

K. Sima'an, 1996. Computational Complexity of Probabilistic Disambiguation by means of Tree Grammars, *Proceedings COLING-96*, Copenhagen, Denmark.

K. Sima'an, 1999. *Learning Efficient Disambiguation*. PhD thesis, ILLC dissertation series number 1999-02. Utrecht/Amsterdam, The Netherlands.

K. Sima'an, 2000. Tree-gram Parsing: Lexical Dependencies and Structural Relations, *Proceedings ACL'2000*, Hong Kong, China.

A. Way, 1999. A Hybrid Architecture for Robust MT using LFG-DOP, *Journal of Experimental and Theoretical Artificial Intelligence* 11 (Special Issue on Memory-Based Language Processing).

R. Weischedel, M. Meteer, R, Schwarz, L. Ramshaw and J. Palmucci, 1993. Coping with Ambiguity and Unknown Words through Probabilistic Models, *Computational Linguistics*, 19(2).